\documentclass{article}

\usepackage{PRIMEarxiv}

\usepackage[utf8]{inputenc} 
\usepackage[T1]{fontenc}    
\usepackage[hidelinks]{hyperref}
\usepackage{url}            
\usepackage{booktabs}       
\usepackage{amsfonts}       
\usepackage{nicefrac}       
\usepackage{microtype}      
\usepackage{lipsum}
\usepackage{fancyhdr}       
\usepackage{graphicx}       
\usepackage{times}
\usepackage{amsmath}
\usepackage{tabularx}
\usepackage{array}
\usepackage{multirow}
\usepackage{makecell}
\usepackage[table]{xcolor}
\usepackage{rotating}
\usepackage{placeins}
\usepackage{afterpage}
\usepackage[numbers]{natbib}
\usepackage{subcaption}
\usepackage{adjustbox}
\usepackage{algorithm}
\usepackage{algpseudocode}

\title{Advancing Interaction-Sensitive Feature Selection: Novel Relief-Based Algorithms, Expanded Comparisons, and Recommendations for Biomedical Data Mining
}

\author{
  Kia Kazemi-Nia, Harsh Bandhey, Philip J. Freda, Ryan J. Urbanowicz\thanks{\hangindent=1.8em Corresponding author\\\textit{Email address}: ryan.urbanowicz@csmc.edu} \\
  Cedars-Sinai Health Sciences University, Los Angeles, 90048, CA, USA \\
}

\begin{document}
\maketitle

\begin{abstract}

\noindent\textbf{Objective:} As a precursor to high-dimensional biomedical data modeling, reliable feature selection can reduce computational expense, improve modeling performance, and yield simpler, more interpretable models. However, most filter-based feature selection methods struggle to detect feature interactions (\textit{e.g.} epistasis), while wrapper or embedded feature selection methods are computationally expensive or overly algorithm-specific. \textit{Relief-based algorithms} (RBAs) are filter methods that are sensitive to feature interactions while mitigating these other limitations. This study (1) refactors, optimizes, and expands the \textit{scikit-rebate} Python package with existing and newly proposed RBA variants and (2) conducts rigorous RBA benchmark comparisons across diverse genomic simulations.

\noindent\textbf{Methods:} We expand \textit{scikit-rebate} to include \textit{SWRF*}, \textit{$\mu$-Relief}, and 5 novel RBA variants implementing alternative strategies for neighbor selection and feature scoring. All RBAs were evaluated to compare predictive feature ranking and runtime across simulated genomic datasets varying in sample size, number of features, heritability, and underlying association type (\textit{e.g.} main effects and interactions).

\noindent\textbf{Results:} All RBAs, except $\mu$-Relief, were proficient in detecting 2-way interactions in noisy data. RBAs utilizing `far' scoring were best at detecting 2-way interactions -- with \textit{MultiSWRFDB*} top-performing -- but were far less sensitive to main effects. \textit{SWRF}, \textit{MultiSWRF}, \textit{MultiSURF}, and \textit{MultiSWRFDB} yielded top performance across main effect and 2-way interaction datasets with \textit{MultiSWRFDB} performing best when also considering 3-way interactions. Furthermore, refactoring of \textit{scikit-rebate} resulted in 10 to 35-fold reductions in RBA runtimes. 

\noindent\textbf{Conclusion:} The improved \textit{scikit-rebate} package provides a highly effective and efficient framework for interaction-sensitive feature selection. The newly introduced algorithms are among the strongest performing, and by robustly retaining both univariate (main) effects and 2-way epistatic interactions, these algorithms preserve predictive signals for downstream modeling.

\end{abstract}


\section{Introduction}\label{intro}
Feature selection is an essential part of many data mining and machine learning pipelines. Its goal is to identify and select the most relevant and informative features from a dataset to construct simpler, better performing, and less overfit models \cite{chandrashekar2014survey,li2017featureselection,cai2018featureselection}. It serves as a form of dimensionality reduction that preserves original feature meaning rather than transforming or projecting data into a new mathematical space (\textit{e.g.} principle component analysis or autoencoders) \cite{GuyonElisseeff2003}. 

A variety of feature selection algorithms have been proposed that fall into either the \textit{filter}, \textit{wrapper}, or \textit{embedded} method categories. Given that wrapper and embedded methods are both algorithm-dependent, meaning that selected features may not ideally transfer from one algorithm/model to another, and that wrapper methods are extremely computationally expensive, `filter' methods are particularly appealing \cite{Relief_review_paper}. Filter methods more rapidly evaluate the relevance of features independently of any machine learning algorithm, instead relying on intrinsic statistical properties of the data to score and rank features for selection. While different feature selection methods have demonstrated domain-specific strengths, very few have been able to consistently and efficiently detect features whose main source of predictive value comes from interactions \cite{zhao2009searching,Relief_review_paper}. Feature selection algorithms that can do this are either computationally expensive, such as FOCUS which exhaustively searches over all subsets of the given feature set \cite{FOCUSpaper}, are wrapper methods with the aforementioned limitations \cite{GuyonElisseeff2003, BolonCanedo2013, Jovic2015}, or are filter methods with a limited ability to detect interactions without main effects to guide them, \textit{e.g.} joint mutual information \cite{yang1999data,brown2012conditional} and correlation-based feature selection \cite{hall1999correlation}.

\textit{Relief-based algorithms} (RBAs) are filter-based feature selection methods developed to address these limitations. Derived from the original Relief algorithm \cite{kira1992feature,kira1992practical}, RBAs leverage local neighborhood information by evaluating target instances against their neighbor instances, defined by feature space proximity, to identify features that distinguish between different outcome values. RBAs can detect pure epistasis (\textit{i.e.}, associations with no `marginal'/ `main' effects) in a computationally efficient manner (since they do not exhaustively explore feature combinations) and they are not dependent on any one induction algorithm as a wrapper. Specifically, RBAs scale linearly with feature count. While they scale quadratically with number of instances, many biomedical datasets have limited sample sizes, and those with large instance counts can be subsampled for RBA scoring efficiency \cite{Relief_review_paper}. This scalability and sensitivity have driven their widespread use in genetics \cite{BerettaSantaniello2011,zhang2008geneticsapplication,sun2019hybridgeneticsapplication}, medical imaging \cite{SahaNandi2024,jasti2022medicalimagingapplication,lu2017medicalimagingapplication}, and other fields where high feature dimensionality is a concern and features involved in interactions are of interest. A comprehensive introduction and review of fundamental RBA mechanics and algorithm variants was published in 2018 \cite{Relief_review_paper}.

Urbanowicz \textit{et al.} \cite{Urbanowicz2018Benchmarking} provided one of the first comprehensive benchmark comparisons of `core' RBAs, and introduced \textit{scikit-rebate} as an open-source Python package implementing these methods for use in data with categorical or quantitative features/outcomes as well as missing values. Core RBAs are defined here as RBA variants that perform a single pass over the training data \cite{Urbanowicz2018Benchmarking}. These are distinct from RBA-wrapper strategies such as TuRF \cite{moore2007tuning} or VLSRelief \cite{eppstein2008very} which apply core RBAs multiple times towards improving performance on large feature sets. While all core RBAs share fundamental neighborhood-based scoring mechanisms, specific variants diverge in their criteria for neighborhood inclusion and weighting of instances in the neighborhood. Variants compared in the aforementioned study \cite{Urbanowicz2018Benchmarking} included ReliefF \cite{ReliefFpaper}, SURF \cite{Greene2009SURF}, SURF* \cite{Greene2010SURFstar},  MultiSURF* \cite{MultiSURFpaper}, and MultiSURF \cite{Urbanowicz2018Benchmarking}. Performance of these RBAs was contrasted with univariate and wrapper approaches including $\chi^2$, ANOVA F-value, Mutual Information \cite{hoque2014mifs}, and ExtraTrees \cite{geurts2006extremely}. The primary findings of that study (1) demonstrated that RBAs using `far' scoring (denoted by a `*' in RBA naming) under-perform in detecting main effects and 3-way interactions, (2) identified MultiSURF as the best general-use RBA, and (3) identified MultiSURF* \cite{MultiSURFpaper} as best for detecting pure 2-way interactions. 

The present study addresses key limitations of the previous benchmarking study \cite{Urbanowicz2018Benchmarking}. First, many of the simulated datasets included only 20 features, leading to some misleading speculations regarding potential performance on higher-order epistasis. Similar benchmarking later conducted for ReliefF, MultiSURF, and MultiSURF* in 100 feature datasets more definitively revealed the inability of RBAs to detect pure 4 or 5-way epistasis \cite{Freda2024}. Additionally, benchmarking had not included gradient-based algorithms like SWRF* \cite{SWRFpaper} or other recently-proposed variants like $\mu$-Relief \cite{MuReliefpaper}. While the studies that introduced $\mu$-Relief \cite{MuReliefpaper} and SWRF* \cite{SWRFpaper} included comparative analyses against other RBAs, their benchmarking methodologies had notable limitations. Specifically, evaluation of SWRF* lacked a comprehensive set of RBA comparisons and failed to assess performance on main effects \cite{SWRFpaper}. Conversely, the $\mu$-Relief study utilized datasets with insufficient interaction effects and indirectly relied on a restrictive set of downstream predictive models for RBA evaluation rather than feature ranking in controlled simulations \cite{MuReliefpaper}. 

This study expands and refactors the \textit{scikit-rebate} package to (1) improve algorithm runtime efficiency, (2) add SWRF* \cite{SWRFpaper} and $\mu$-Relief \cite{MuReliefpaper}, and (3) introduce 5 novel core RBA variants that recombine neighborhood and scoring strategies from SWRF* \cite{SWRFpaper} and MultiSURF \cite{Urbanowicz2018Benchmarking}, namely SWRF, MultiSWRF, MultiSWRF*, MultiSWRFDB, and MultiSWRFDB*. We evaluate the strengths and weaknesses of each algorithm across a diversity of simulated genomic datasets and provide updated recommendations for RBA application in future biomedical data mining tasks.







\section{Methods}\label{methods}

This section covers (1) existing RBAs, (2) newly proposed RBAs, (3) updates to the \textit{scikit-rebate} Python package, (4) benchmarking datasets, (5) experimental analyses, and (6) post-hoc analyses.

\subsection{Existing Relief-based Algorithms}\label{subsec:core_rbas_assessed}

All RBAs are descendants of the original Relief algorithm \cite{kira1992feature,kira1992practical} and return weights ($W$) for all features after training, where higher weights denote greater feature importance. As detailed in Algorithm \ref{alg:relief}, the original Relief algorithm was designed for classification data and operates by iterating over $m$ training instances, each taking a turn as the `target' instance ($R_i$). Each iteration, $R_i$ is compared to its nearest hit ($H$) and nearest miss ($M$), \textit{i.e.}, \textit{neighboring} instances with the \textit{same} or \textit{different} class, respectively. Weights are then updated for each feature ($W[A]$). Feature value differences between $R_i$ and H contribute to weight reductions, and feature value differences between $R_i$ and M increase weights. A key to Relief feature interaction sensitivity is that weight updates take place only among neighboring instances. 

\begin{algorithm}[h]
\caption{Relief algorithm pseudo-code}
\label{alg:relief}
\begin{algorithmic}

\Require For each training instance: a vector of feature values and the class value
\State $n \gets$ number of training instances
\State $a \gets$ number of features
\State \textbf{Hyperparameter}: $m \gets$ number of random training instances out of $n$ used to update $W$
\Statex
\State Initialize all feature weights $W[A] := 0.0$

\For{$i := 1$ to $m$}
    \State Randomly select a target instance $R_i$
    \State Find nearest hit $H$ and nearest miss $M$
    
    \For{$A := 1$ to $a$}
        \State $W[A] := W[A]
        - \frac{\mathrm{diff}(A,R_i,H)}{m}
        + \frac{\mathrm{diff}(A,R_i,M)}{m}$
    \EndFor
\EndFor

\State \Return the vector $W$ of feature scores that estimate the quality of features

\end{algorithmic}
\end{algorithm}

\begin{figure*}[t]
    \centering
    \includegraphics[width=0.9\textwidth]{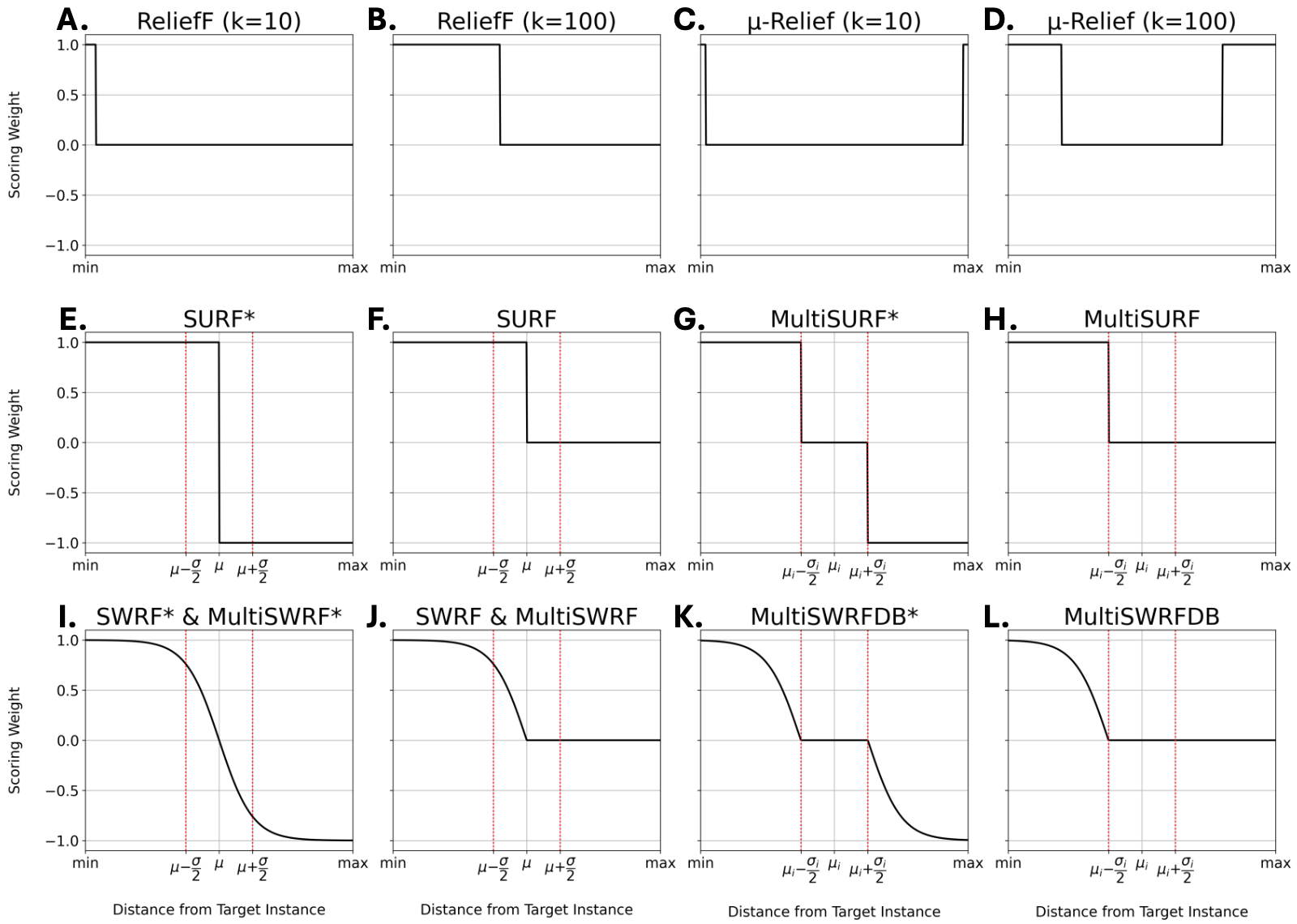}
    \caption{Illustration of instance neighborhoods and scoring weights ($w_{ij}$) based on distance from target instance ($R_i$) for all RBAs examined. Conceptual visualizations generated with 500 simulated instances. While not indicated in subplots I and J, MultiSWRF* and MultiSWRF recalculate $\mu_i$ and $\sigma_i$ for each $R_i$, like other `Multi' RBAs.}
    \label{fig:instanceweight_plots}
\end{figure*}

Since Relief, a multitude of RBA variants have been proposed which introduced novel algorithmic elements aimed at improving sensitivity to relevant features \cite{Relief_review_paper}. These include (1) using more neighbors in scoring (ReliefF \cite{ReliefFpaper}), (2) automatically defining the neighborhood based on inherent dataset properties (SURF \cite{Greene2009SURF}), (3) adding `far' scoring (SURF* \cite{Greene2010SURFstar}), (4) recalculating neighborhood-defining metrics for each $R_i$ (MultiSURF* \cite{MultiSURFpaper}), (5) adding a `deadband (DB) zone' of \textit{middle} distance instances excluded from $W$ updates (MultiSURF* \cite{MultiSURFpaper}), and (6) updates to feature weights $W$ being modulated by neighbors' \textit{scoring weights} ($w_{ij}$) that depend on distance from $R_i$ (SWRF* \cite{SWRFpaper}). Subsections below detail the existing RBAs compared in this study. Of these, $\mu$-Relief and SWRF* are newly implemented in \textit{scikit-rebate}. Figure \ref{fig:instanceweight_plots} is used throughout to illustrate conceptual differences in RBA instance neighborhoods and scoring weights. RBAs are introduced chronologically in sections below, but grouped by scoring weight similarity in Figure \ref{fig:instanceweight_plots}. For a more detailed overview and review of RBAs, see \cite{Relief_review_paper} and \cite{Urbanowicz2018Benchmarking}.

\subsubsection{ReliefF}\label{ReliefFintro}
ReliefF \cite{ReliefFpaper} is arguably the best known descendant of Relief \cite{kira1992feature,kira1992practical}. ReliefF adds the hyperparameter \emph{k}, which defines the number of nearest hits and nearest misses to utilize when updating $W$. Original Relief would have \emph{k}=1. Using a larger \emph{k} can improve algorithm sensitivity to relevant features in noisy data \cite{ReliefFpaper}, however setting \emph{k} too high can dramatically reduce sensitivity to interactions in datasets with small training sets \cite{Urbanowicz2018Benchmarking}. Figures \ref{fig:instanceweight_plots}-A and \ref{fig:instanceweight_plots}-B illustrate the scoring neighborhood and \textit{scoring weights} ($w_{ij}$) used by ReliefF for \emph{k}=10 and \emph{k}=100, respectively. All instances included in the neighborhood receive `standard' scoring weights, ($w_{ij}=1$), meaning that feature value differences with `hits' reduce $W[A]$ by $1/m$ and increase by $1/m$ for `misses'. Instances outside the neighborhood have $w_{ij}=0$, \textit{i.e.}, they don't participate in updating $W$.

\subsubsection{SURF}
SURF \cite{Greene2009SURF} eliminated the need for users to specify ReliefF's \emph{k} hyperparameter; instead automatically identifying \textit{nearest} neighbors based on a one-time calculation of global mean distance ($\mu$) between all pairs of instances in the training data. Here, neighbors for a specific $R_i$ include other instances whose distance from $R_i$ is less than $\mu$. Figure \ref{fig:instanceweight_plots}-F illustrates this larger instance neighborhood, again using `standard' scoring weights.

\subsubsection{SURF*}
SURF* \cite{Greene2010SURFstar} was the first RBA to introduce a `far' neighborhood. For each $R_i$, all instances in the training data are separated into respective \textit{near} and \textit{far} neighborhoods using SURF's global mean distance ($\mu$) \cite{Greene2009SURF}. Figure \ref{fig:instanceweight_plots}-E illustrates how instances in the \textit{far} neighborhood have an inverted scoring weight (\textit{i.e.}, $-1$), meaning that feature value differences with `hits' now \textit{increase} $W[A]$ by $1/m$ and \textit{decrease} by $1/m$ for `misses'. While \textit{far} scoring provides an advantage in the detection of 2-way interactions \cite{Greene2010SURFstar,Urbanowicz2018Benchmarking} it also reduces sensitivity to main effects \cite{Urbanowicz2018Benchmarking}. RBAs that utilize \textit{far} neighborhoods in feature weighting are denoted with `*' in their names.

\subsubsection{SWRF*}
Rather than assigning scoring weights ($w_{ij}$) of 1 or -1 to near/far neighbors like SURF*, SWRF* \cite{SWRFpaper} proposed a sigmoidal gradient to smoothly decrease neighbors' $w_{ij}$ as distance from $R_i$ approaches $\mu$ (Figure \ref{fig:instanceweight_plots}-I). Equation \ref{eq:swrfstar_weight} gives the calculation of $w_{ij}$ based on the distance ($dist_{ij}$) between $R_i$ and another instance $R_j$. The global mean distance, $\mu$, and global standard deviation, $\sigma$, are calculated from all pairwise distances, similar to SURF and SURF*.

{\small
\begin{equation}
w_{ij} = \frac{2}{1 + \exp\left( ( \mathrm{dist}_{ij} - \mu ) \cdot \frac{4}{\sigma} \right)} - 1
\label{eq:swrfstar_weight}
\end{equation}
}

\subsubsection{MultiSURF*}
MultiSURF* \cite{MultiSURFpaper} separately expanded upon SURF* with two key elements: recalculation of neighborhood metrics for each $R_i$ (\textit{i.e.}, the prefix `Multi') and a `DB zone'. Unlike SURF*, the metrics that determine neighborhood membership in MultiSURF*, including $\mu$, are recalculated for each $R_i$ (\textit{i.e}., $\mu_i$) based on distances between $R_i$ and all other instances. The DB zone is defined by calculating the standard deviation ($\sigma_i$) of these distances from $R_i$. Here the \textit{near} neighborhood is defined by instances with a distance from $R_i$ that is $<$ ($\mu_i - \frac{\sigma_i}{2}$) and the \textit{far} neighborhood includes instances $>$ ($\mu_i + \frac{\sigma_i}{2}$). As seen in Figure \ref{fig:instanceweight_plots}-G, instances within the DB zone receive $w_{ij}=0$.

\subsubsection{MultiSURF}
MultiSURF \cite{Urbanowicz2018Benchmarking} is identical to MultiSURF*, but importantly \textit{excludes} `far' neighbors in feature scoring (Figure \ref{fig:instanceweight_plots}-H). Comprehensive benchmarking suggested that MultiSURF was the best general-use RBA for detecting main effects, 2-way and 3-way interactions, and genetic heterogeneity \cite{Urbanowicz2018Benchmarking,Freda2024}. However, SWRF* was not included in those comparisons, as it was expected that SWRF*'s `far' scoring would lead to a similar loss in main effect sensitivity as observed in \cite{Urbanowicz2018Benchmarking}. This is examined in the current study.

\subsubsection{\boldmath$\mu$-Relief}
$\mu$-Relief (Mean-based Relief), \cite{MuReliefpaper} was most recently proposed as a variant of ReliefF \cite{ReliefFpaper} that redefined `neighborhoods' to be the \emph{k} `hit' and \emph{k} `miss' instances with the largest $\mathrm{dist}_{ij}^{*}$ (Equation \ref{eq:murelief_dist}). This aims to maximize the absolute difference between $\mathrm{dist}_{ij}$ and the average distance between $R_i$ and the subset of instances with the same class as $R_j$ (\textit{i.e.}, $\mu_{C_j i}$). 

{\small
\begin{equation}
\mathrm{dist}_{ij}^{*} = \left| \mathrm{dist}_{ij} - \mu_{C_j i} \right|
\label{eq:murelief_dist}
\end{equation}
}

$\mu$-Relief's neighborhood selection inevitably leads to the inclusion of `far neighbors', similar to other `star' RBAs, \textit{e.g.} SURF* \cite{Greene2010SURFstar}. However, unlike `star' RBAs that assign negative scoring weights to the `far neighborhood', $\mu$-Relief maintains `standard' \textit{scoring weights} for all instances in the neighborhood as illustrated in Figure \ref{fig:instanceweight_plots}-C and \ref{fig:instanceweight_plots}-D, for \emph{k}=10 and \emph{k}=100, respectively. No $\mu$-Relief implementation was publicly available, so the present study implements it as described in \cite{MuReliefpaper} and conducts the first rigorous simulation study benchmarking in comparison with other RBAs. Lastly, while original $\mu$-Relief was only designed for binary and multi-class outcomes, our implementation in \textit{scikit-rebate} was extended for quantitative outcomes as described in \cite{Urbanowicz2018Benchmarking}.

\subsection{Novel RBA Variants}\label{subsec:novel_rbas_assessed}
This section describes the novel RBA variants that were implemented and compared in this study.

\subsubsection{MultiSWRF*}
MultiSWRF* combines (1) SWRF*'s sigmoidal gradient for $w_{ij}$ with (2) MultiSURF*'s recalculation of $\mu_i$ and $\sigma_i$ for each $R_i$, rather than a single global metric calculation (Equation \ref{eq:multiswrfstar_weight} and Figure \ref{fig:instanceweight_plots}-I). SWRF* and MultiSWRF* are both represented by Figure \ref{fig:instanceweight_plots}-I since their only difference is $\mu$ and $\sigma$ calculation. Like SWRF*, MultiSWRF* speculates that it is better for all instances to participate in updating $W$. In contrast with SWRF*, it examines whether recalculated neighborhood metrics improve performance.

{\small
\begin{equation}
w_{ij} = \frac{2}{1 + \exp\left( ( \mathrm{dist}_{ij} - \mu_i ) \cdot \frac{4}{\sigma_i} \right)} - 1
\label{eq:multiswrfstar_weight}
\end{equation}
}

\subsubsection{SWRF}
SWRF utilizes the same sigmoidal gradient and global $\mu$ and $\sigma$ as SWRF* \cite{SWRFpaper} but it \textit{excludes} `far' neighbors in feature scoring (Equation \ref{eq:swrf_weight} and Figure \ref{fig:instanceweight_plots}-J). This is expected to recover performance on main effects similar to MultiSURF vs. MultiSURF* \cite{Urbanowicz2018Benchmarking}.

{\small
\begin{equation}
w_{ij} =
\begin{cases}
\displaystyle
\frac{2}{1 + \exp\left( (\mathrm{dist}_{ij} - \mu)\cdot \frac{4}{\sigma} \right)} - 1
& \text{if } \mathrm{dist}_{ij} < \mu \\[10pt]

0
& \text{otherwise}
\end{cases}
\label{eq:swrf_weight}
\end{equation}
}

\subsubsection{MultiSWRF}
MultiSWRF is identical to SWRF but adopts recalculation of $\mu_i$ and $\sigma_i$ for each $R_i$, like MultiSWRF* (Equation \ref{eq:multiswrf_weight} and Figure \ref{fig:instanceweight_plots}-J). As a non-`*' algorithm, we expect better performance on main effects.

{\small
\begin{equation}
w_{ij} =
\begin{cases}
\displaystyle
\frac{2}{1 + \exp\left( (\mathrm{dist}_{ij} - \mu_i)\cdot \frac{4}{\sigma_i} \right)} - 1
& \text{if } \mathrm{dist}_{ij} < \mu_i \\[10pt]

0
& \text{otherwise}
\end{cases}
\label{eq:multiswrf_weight}
\end{equation}
}

\subsubsection{MultiSWRFDB*}
MultiSWRFDB* is similar to MultiSWRF* but also includes the DB zone introduced by MultiSURF* \cite{MultiSURFpaper} (Figure \ref{fig:instanceweight_plots}-K and Equation \ref{eq:multiswrfdbstar_weight}). In contrast with MultiSWRF*, MultiSWRFDB* speculates that `middle' distance instances, falling within the DB, should be excluded from $W$ updates (\textit{i.e.}, $w_{ij}=0$).

{\footnotesize
\begin{equation}
w_{ij} =
\begin{cases}
\displaystyle
\frac{2}{1 + \exp\!\left( (\mathrm{dist}_{ij} - (\mu_i - \frac{\sigma_i}{2})) 
\cdot \frac{4}{\sigma_i} \right)} - 1,
& \text{if } \mathrm{dist}_{ij} < \mu_i - \frac{\sigma_i}{2}
\\[10pt]

\displaystyle
\frac{-2}{1 + \exp\!\left( (\mathrm{dist}_{ij} - (\mu_i + \frac{\sigma_i}{2})) 
\cdot \frac{-4}{\sigma_i} \right)} + 1,
& \text{if } \mathrm{dist}_{ij} > \mu_i + \frac{\sigma_i}{2}
\\[10pt]

0
& \text{otherwise}
\end{cases}
\label{eq:multiswrfdbstar_weight}
\end{equation}
}

\subsubsection{MultiSWRFDB}
MultiSWRFDB is identical to MultiSWRFDB*, but it \textit{excludes} `far' neighbors in feature scoring (Figure \ref{fig:instanceweight_plots}-L and Equation \ref{eq:multiswrfdb_weight}). We hypothesize that MultiSWRFDB will perform best overall given that it combines elements from existing RBAs that have proven to be advantageous in various contexts, \textit{i.e.}, (1) excluding `far' neighbors in feature scoring, (2) sigmoidal gradient for $w_{ij}$, (3) recalculation of $\mu_i$ and $\sigma_i$ for each $R_i$, and (4) instances within the DB zone having $w_{ij}=0$.

{\footnotesize
\begin{equation}
w_{ij} =
\begin{cases}
\displaystyle
\frac{2}{1 + \exp\!\left( (\mathrm{dist}_{ij} - (\mu_i - \frac{\sigma_i}{2})) 
\cdot \frac{4}{\sigma_i} \right)} - 1,
& \text{if } \mathrm{dist}_{ij} < \mu_i - \frac{\sigma_i}{2}
\\[10pt]

0
& \text{otherwise}
\end{cases}
\label{eq:multiswrfdb_weight}
\end{equation}
}

\subsection{Updates to \textit{scikit-rebate} Python package}

All core RBAs described in Section \ref{subsec:core_rbas_assessed} have been implemented in the scikit-learn compatible \textit{scikit-rebate} Python package which is publicly available at \url{https://github.com/UrbsLab/scikit-rebate}. We also introduce efficiency and utility updates to \textit{scikit-rebate} including: (1) refactoring of the codebase to achieve significant speedups in feature scoring via NumPy vector-wide operations, and (2) explicit user control of how individual features and endpoints are treated, \textit{i.e.}, as categorical or quantitative values. These expansions are reflected in v0.8.3 of \textit{scikit-rebate} which was employed for all subsequent evaluations in this study. Furthermore, we have corrected the equations describing feature weight updates for multi-class endpoints (Eqs. 4-5 in \cite{Urbanowicz2018Benchmarking}) to accurately reflect these updates within \textit{scikit-rebate} (Supplementary Section 1.1).

\subsection{Simulated Benchmark Datasets}
In-line with other rigorous RBA feature ranking evaluations \cite{Urbanowicz2018Benchmarking,Freda2024}, this study benchmarks all aforementioned RBAs across a diverse spectrum of simulated genetic and `XOR' datasets, generated using the GAMETES v2.2 software package \cite{urbanowicz2012gametes} and custom scripts. Simulation studies are critical for algorithm benchmarking as they allow systematic control over experimental dataset conditions, and the ground-truth of the dataset is known to investigators including which features were modeled as relevant or irrelevant, the signal-to-noise ratio, and the nature of associations in the data. 

As summarized in Table \ref{tab:datasets_used}, we examine 77 dataset configurations that vary by (1) the type of association between predictive feature(s) and outcome (\textit{e.g.} main, additive, epistatic, and/or heterogeneous effects), (2) number of predictive features, (3) total number of features, (4) number of instances, (5) heritability (\textit{i.e.}, the signal-to-noise ratio, where 1 is full penetrance and 0 is complete noise), (6) model `architecture' difficulty; `E' (easy) vs. `H' (hard) \cite{urbanowicz2012predicting}, and (7) various configuration-specific variables detailed in \cite{Urbanowicz2018Benchmarking}, \textit{e.g.} XOR 2-way, 3-way, 4-way, or 5-way interactions. This study primarily focuses on the first 63 dataset configurations, but secondarily confirms RBA functionality on smaller (20-feature) datasets with continuous-valued features, continuous-valued endpoints, missing values, class imbalance, and multi-class endpoints. Each `configuration' includes 30 replicate datasets used to evaluate the success rate of RBA feature ranking across multiple trials, for a total of 2310 unique simulated datasets. Like more recent benchmarking in \cite{Freda2024}, primary analyses were conducted on datasets with $100$ or more features. We largely focus on pure epistasis since it is more challenging to detect than `impure' epistasis, where small marginal effects can facilitate detection.  

\begin{table*}[t]
\centering
\caption{Summary of simulation study benchmark datasets. The simulation method is designated as ‘G’ (GAMETES), 
‘C’ (custom script), or ‘G + C’ (GAMETES modified by custom script). Primary data configurations are given in the top portion of the table, and secondary ones on the bottom.}
\label{tab:datasets_used}
\small
\renewcommand{\arraystretch}{1.2}

\resizebox{\linewidth}{!}{%
\begin{tabular}{%
>{\raggedright\arraybackslash}p{4cm}
c
c
c
>{\centering\arraybackslash}p{4cm}
c
c
c
c}
\toprule

\textbf{Simulated Data Group Description or Pattern of Association} \textit{(`*' indicates datasets were also simulated with 2-way pure epistasis)} &
\raisebox{-10ex}{\rotatebox{90}{\textbf{Configurations}}} &
\raisebox{-10ex}{\rotatebox{90}{\textbf{Config. Variations}}} &
\raisebox{-10ex}{\rotatebox{90}{\textbf{Predictive Features}}} &
\raisebox{-10ex}{\rotatebox{90}{\textbf{Total Features}}} &
\raisebox{-10ex}{\rotatebox{90}{\textbf{Model Difficulty}}} &
\raisebox{-10ex}{\rotatebox{90}{\textbf{Heritability}}} &
\raisebox{-10ex}{\rotatebox{90}{\textbf{Instances}}} &
\raisebox{-10ex}{\rotatebox{90}{\textbf{Simulation Method}}} \\

\midrule

\rowcolor{gray!15}
2-way Pure Epistasis 
& 32 & -- & 2 & 100 & E, H 
& 0.05, 0.1, 0.2, 0.4 
& 200, 400, 800, 1600 
& G \\

XOR Model (Pure Epistasis)
& 4 & 2-way, 3-way, 4-way, 5-way 
& 2, 3, 4, 5 
& 100 & N/A 
& 1 & 1600 & C \\

\rowcolor{gray!15}
4-Feat. Heterogeneous 2-way Epistasis
& 2 & 50:50, 75:25 & 4 & 100 & E 
& 0.4 & 1600 & G \\

3-way Pure Epistasis
& 1 & -- & 3 & 100 & E 
& 0.2 & 1600 & G \\

\rowcolor{gray!15}
1-Feature Main Effect
& 8 & -- & 1 & 100 & E, H 
& 0.05, 0.1, 0.2, 0.4 
& 1600 
& G \\

2-Feature Additive Effect
& 2 & 50:50, 75:25 & 2 & 100 & E 
& 0.4 & 1600 & G \\

\rowcolor{gray!15}
4-Feature Additive Effect
& 1 & -- & 4 & 100 & E 
& 0.4 & 1600 & G \\

Number of Features*
& 9 & 2-way Pure Epistasis & 2 
& 100, 1000, 2000, 5000, 8000, 10000, 20000, 50000, 100000 
& E & 0.4 & 1600 & G \\

\rowcolor{gray!15}
Number of Features 
& 4 & Main Effect & 1 
& 100, 1000, 10000, 100000 
& E & 0.4 & 1600 & G \\ \hline

Continuous Features*
& 1 & -- & 2 & 20 & E 
& 0.4 & 1600 & G+C \\

\rowcolor{gray!15}
Mix of Discrete and Continuous Features*
& 1 & -- & 2 & 20 & E 
& 0.4 & 1600 & G+C \\

Continuous Endpoint*
& 3 & 0.2, 0.5, 0.8 & 2 & 20 & E 
& 0.4 & 1600 & G \\

\rowcolor{gray!15}
Continuous Endpoint* (1-Threshold Model)
& 1 & -- & 2 & 20 & E 
& 0.4 & 1600 & G+C \\

Missing Data*
& 4 & 0.001, 0.01, 0.1, 0.5 
& 2 & 20 & E 
& 0.4 & 1600 & G+C \\

\rowcolor{gray!15}
Imbalanced Data*
& 2 & 0.6, 0.9 & 2 & 20 & E 
& 0.4 & 1600 & G \\

Multi-Class Endpoint
& 2 & 3-class, 9-class 
& 2 & 20 & N/A 
& 1 & 1600 & C \\

(Impure 2-way Epistasis) & & & & & & & & \\

\bottomrule
\end{tabular}
}
\end{table*}

\subsection{Experimental Evaluation}
This study compares the performance of 7 existing RBAs, 5 novel RBAs, Mutual Information, and a random ordering of features (Random Shuffle), as a negative control. Mutual Information is an established feature selection algorithm best suited for detecting main/additive effects \cite{thomas2006information,pedregosa2011scikit}. For ReliefF and $\mu$-Relief, settings of \emph{k}=10 and \emph{k}=100 were examined for each, yielding a total of 14 RBA algorithms being run. 

This simulation study design allows us to directly compare feature selection algorithm performance based on how highly they score/rank predictive features in contrast with non-predictive features. Ideal algorithm performance would score all predictive features higher than all non-predictive features in a given dataset. This design avoids the need for training/testing data splits or indirect evaluation via downstream modeling since we evaluate the accuracy of ground truth feature ranking directly. 

From the perspective of feature selection in real-world applications, practitioners would want to ensure that all predictive features are preserved within the top `selected' subset of $z$ features for downstream machine learning modeling. Thus, this study compares algorithm performance by examining how often (\textit{i.e.}, out of 30 replicates) all predictive features were ranked in the top \emph{x}\% of features. This is examined for all percentiles (\emph{x}\%); from \emph{x=}($\frac{\text{number of predictive features}}{\text{number of total features}} \times 100$)\% (\textit{i.e.}, `optimal' -- all predictive features ranked ahead of non-predictive ones) to \emph{x=100}\% (\textit{i.e.}, predictive features could have any ranking). 

\begin{figure}[t]
\centering
\includegraphics[width=\columnwidth]{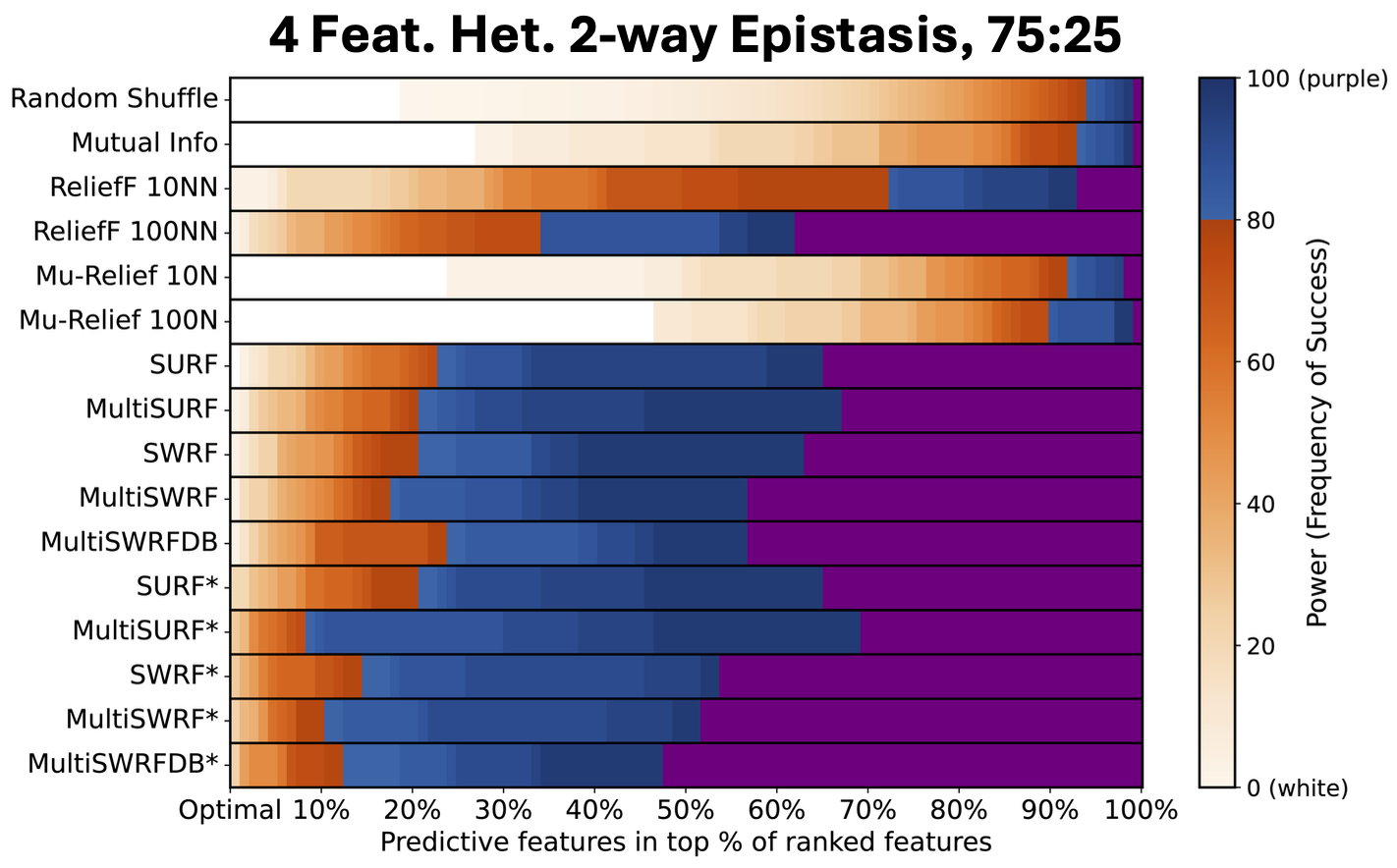}
\caption{Example heatmap illustrating algorithms' ability to rank predictive features in the top \emph{x}\% of features across replicates. This heatmap is for the configuration in the `4-Feature Heterogeneous 2-way Epistasis' data group with a 75:25 ratio (\textit{i.e.}, one 2-way interaction is modeled in 75\% of the instances and a different 2-way interaction is modeled in the other 25\% of instances).}
\label{fig:heatmap_example}
\end{figure}

These rankings were used to generate heatmap visualizations for each dataset configuration, \textit{e.g.} Figure \ref{fig:heatmap_example}. Heatmap \textit{x}-axes represents feature percentiles (\emph{x}\%) and the \textit{y}-axes indicates the algorithm. The legend denotes `power', \textit{i.e.}, the frequency of success across 30 replicates wherein all predictive features were ranked at or above a given percentile. Purple denotes ideal power (\textit{i.e.}, 30/30 replicates), while shades of blue denote power $>80\%$, and shades from orange to white highlight diminishing power. As such, heatmaps represent ideal algorithm performance as a solid purple horizontal bar (\textit{i.e.}, all predictive features ranked $>$ non-predictive features in every replicate). In addition to these individual heatmaps, `heatmap grid' visualizations were also generated which arrange individual dataset-configuration-level heatmaps within a data group into one plot.

\subsubsection{Algorithm Training}
Each RBA was trained on all instances of each dataset to yield respective feature importance scores. \textit{Scikit-rebate} was directed to treat all features as categorical, except for secondary simulated datasets including continuous-valued features. After training, feature scores were sorted in descending order for rank analysis. While RBA scoring was deterministic, a fixed random seed was used for Random Shuffle and Mutual Information to ensure reproducibility.  For datasets with feature counts of 100, 1,000, 10,000, and 100,000, the runtimes of the RBAs were recorded to quantify the computational speedup achieved by \textit{scikit-rebate} refactoring. 
\subsubsection{Post-hoc Analysis}
In addition to predictive feature percentile rankings used in previous benchmarking \cite{Urbanowicz2018Benchmarking,Freda2024}, this study also examined the mean and median rankings given to predictive features which were calculated for each algorithm at the dataset configuration and group levels. These metrics were computed as concise aggregate summary statistics that complement presented heatmap visualizations.

Furthermore, Mann-Whitney U tests were performed to compare algorithm rankings at both the configuration and data group level (\textit{e.g.} 2-way pure epistasis). The tests were performed for all pairwise combinations of algorithms (excluding Random Shuffle). Thus, there were $\binom{15}{2}$ = $105$ tests performed per configuration or data group. Since this ranking data has the potential for many ties/overlapping values between the two compared algorithms which can lead to diminished power in the Mann-Whitney U test, permutation tests were also performed ($n=10,000$ permutations; see Supplementary Section 1.2 for details). To account for multiple hypothesis testing, \textit{p}-values from all Mann-Whitney U tests and permutation tests were adjusted using the Benjamini-Hochberg FDR procedure ($\alpha = 0.05$). Since both tests produced similar \textit{p}-values, the \textit{p}-value threshold that held true for both is reported by default.

Lastly, to provide recommendations on which RBA(s) to employ in distinct scenarios, we computed `global' mean and median ranking metrics. First, all primary data configurations (\textit{i.e.}, those with $\geq$ 100 features) were separated into one of three groups: (1) 2-way epistasis data (44 configurations), (2) additive/main effect data (15 configurations), or (3) 3-way epistasis data (2 configurations). The specific configurations in these groups can be found in Supplementary Section 1.2. 4-way and 5-way XOR datasets were excluded due to poor performance across all algorithms. Next, the group-level mean and median ranking was calculated for each algorithm across all datasets within that group. Ultimately, we reported `global' mean and median metrics for 3 scenarios of expected RBA use: (1) ranking on 2-way epistasis data - for users solely interested in detecting 2-way interactions, (2) ranking on 2-way epistasis and additive/main effect data - for users seeking to detect interactions without sacrificing additive/main effect performance, and (3) ranking for 2-way epistasis, additive/main effect, and 3-way epistasis data - for users seeking the most generalizable RBA capabilities. Rankings on datasets with $>$ 100 features were scaled to a 1-100 range to prevent excessive influence on the mean and median values. For global metrics on the second and third scenarios, group-level mean/median values were averaged. Scripts for conducting all analyses in this paper are available within the \textit{scikit-rebate} GitHub repository.

\section{Results}\label{Results}
This section compares RBAs' rankings of predictive features, organized by (1) 2-way (or greater) epistasis, (2) univariate effects (\textit{i.e.}, main or additive), (3) scaling to larger feature sets, (4) global mean/median metrics for different use scenarios, and (5) secondary datasets.

\subsection{Detecting Feature Interactions}
This section examines RBA performance on pure 2-way, 3-way, 4-way, or 5-way epistasis.

\subsubsection{2-way Epistasis}\label{subsubsec:two_way_interactions}

\begin{figure*}[!h]
\centering
\includegraphics[width=\textwidth]{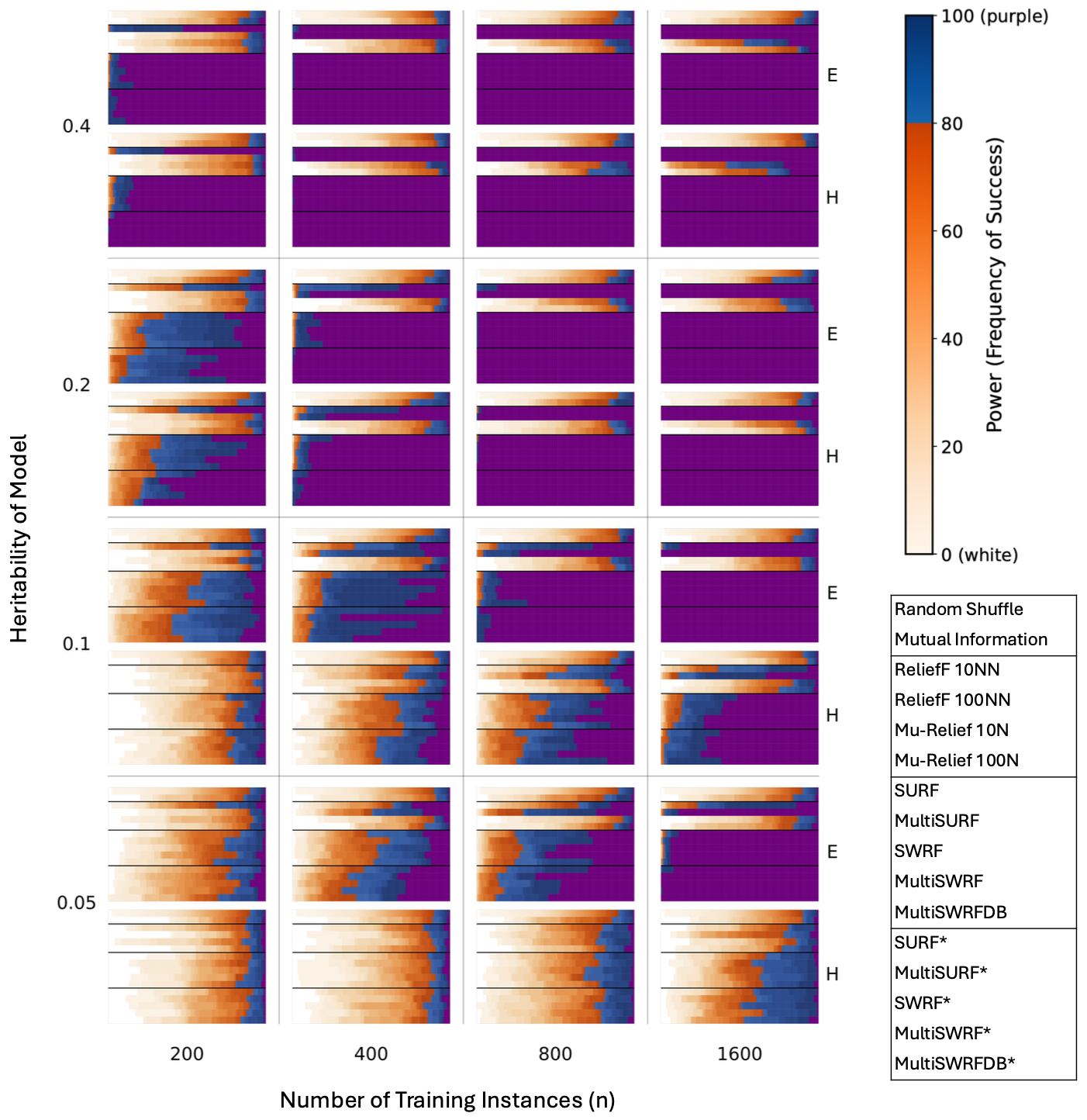}
\caption{Heatmap grid for the `2-way Pure Epistasis' data group. X-axis indicates number of training instances, y-axis (left) indicates heritability, and y-axis (right) indicates model architecture difficulty (easy or hard). Black lines are drawn on individual heatmaps to identify RBA algorithms ordered in the figure key.}
\label{fig:core2wayEpi_unifiedheatmap}
\end{figure*}

The heatmap grid in Figure \ref{fig:core2wayEpi_unifiedheatmap} summarizes RBA performance across the 32 dataset configurations of the `2-way Pure Epistasis' data group. This grid is arranged such that the most difficult configuration (low heritability, low instance count, and hard architecture) is bottom-left, and the easiest configuration is top-right. Overall, most RBAs performed well on these benchmarks, with expected performance losses towards lower extremes of heritability, instance count, and hard architectures. Mean/median rankings given by RBAs (Table S1) reinforce this, with mean values more affected by these more challenging configurations. In contrast, $\mu$-Relief (both \emph{k=10} and \emph{k}=100) completely failed to detect 2-way interactions, with performance comparable to Random Shuffle (negative control) and Mutual Information (expected to perform poorly on pure epistasis).

Consistent with previous benchmarking \cite{Urbanowicz2018Benchmarking}, `star' RBAs (\textit{e.g.} MultiSURF* and MultiSWRFDB*) demonstrated small but overall significant ranking improvements over all `non-star' RBAs ($p<0.01$ to $p<0.0001$; Supplementary File S1). Within either `star' or `non-star' RBAs, those with a `DB zone' (MultiSURF/MultiSURF*, MultiSWRFDB/MultiSWRFDB*) and those utilizing a scoring gradient (SWRF/SWRF*, MultiSWRF/MultiSWRF*) achieved best performance among RBAs, while SURF/SU\discretionary{-}{}{}RF* performed slightly less well, and ReliefF (\emph{k}=10 and \emph{k}=100) performed least well among successful RBAs. The differences between `non-star' DB/gradient algorithms and SURF/ReliefF were significant ($p<0.05$ to $p<0.0001$, except MultiSWRF vs. SURF which was non-significant). The best performance (based on group mean/median) was achieved by MultiSWRFDB*; however, this advantage was small and non-significant compared to all other `star' RBAs, including SURF*.

\begin{figure*}[!h]
\centering
\includegraphics[width=\textwidth]{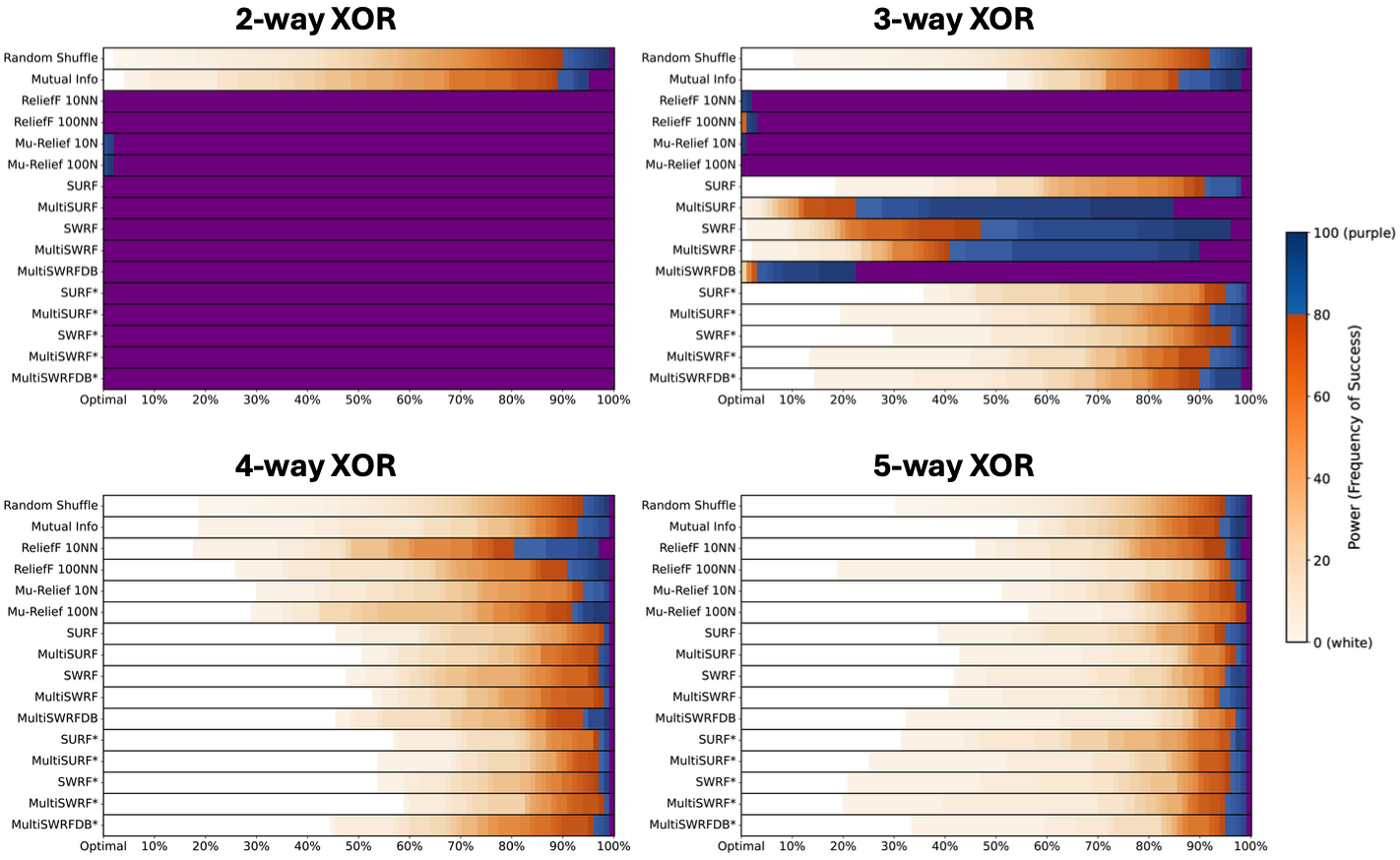}
\caption{Predictive feature ranking heatmaps for the `XOR Model' data group.}
\label{fig:xor_unifiedheatmap}
\end{figure*}

For the easier 2-way XOR configuration (heritability = 1), most RBAs exhibited ideal ranking performance, except $\mu$-Relief which exhibited near-ideal performance (Figure \ref{fig:xor_unifiedheatmap}, Table S2).

On the `4-Feature Heterogeneous 2-way Epistasis' data group, with two separate 2-way interactions modeled in distinct instance subsets, RBA performance differences were very similar to the `2-way Pure Epistasis' data group (see Figure \ref{fig:heatmap_example} for 75:25, Figure S1 for 50:50, and Table S3 for mean/median rankings across the group). Of note, ReliefF (\emph{k}=10) underperformed in contrast with other RBAs. The only significant differences between algorithms involved under-performing $\mu$-Relief (\emph{k}=10 and \emph{k}=100) or Mutual Information ($p<0.0001$; Supplementary File S2). However, here there are far fewer dataset configurations, decreasing overall power of the analysis.

\subsubsection{3-way Epistasis}\label{subsubsec:three_way_interactions}

For the `3-way XOR' configuration (also heritability = 1), RBAs that utilized fewer neighbors in scoring (ReliefF with \emph{k}=10/\emph{k}=100 and $\mu$-Relief with \emph{k}=10/\emph{k}=100) performed the best, achieving ideal or near-ideal performance (Figure \ref{fig:xor_unifiedheatmap}, Table S4). Conversely, SURF and every `star' RBA failed to detect 3-way XOR predictive features. MultiSWRFDB stood out as the only RBA to perform well on the 3-way XOR that also performed well on other dataset configurations. Following MultiSWRFDB; MultiSURF, SWRF and MultiSWRF yielded moderate but diminishing performance.

These findings were generally echoed by the `3-way Pure Epistasis' configuration which included much noisier signal (heritability = 0.2). While overall RBA performance weakened with significant noise, $\mu$-Relief and ReliefF remained the strongest performers, with $\mu$-Relief (\emph{k}=100) achieving highest performance (Figure S2). On both of the above 3-way configurations, $\mu$-Relief (\emph{k}=10) and $\mu$-Relief (\emph{k}=100) had performance improvements that were significant compared to all RBAs, as did ReliefF (\emph{k}=10) and ReliefF (\emph{k}=100) ($p<0.05$ to $p<0.0001$; see Supplementary Files S3 and S4).

Despite failing on 2-way epistasis, $\mu$-Relief was the strongest performer on datasets involving 3-way epistasis, and, in contrast, `star' RBAs, which were the strongest in detecting 2-way epistasis, failed to detect 3-way epistasis.

\subsubsection{4-way and 5-way Epistasis}
Consistent with previous findings \cite{Urbanowicz2018Benchmarking}, no RBAs examined were able to reliably detect features involved in XOR 4-way and 5-way interactions given 1600 training instances (Figure \ref{fig:xor_unifiedheatmap}, Tables S6 and S7).

\subsection{Detecting Univariate Effects}

\begin{figure*}[b]
\centering
\includegraphics[width=\textwidth]{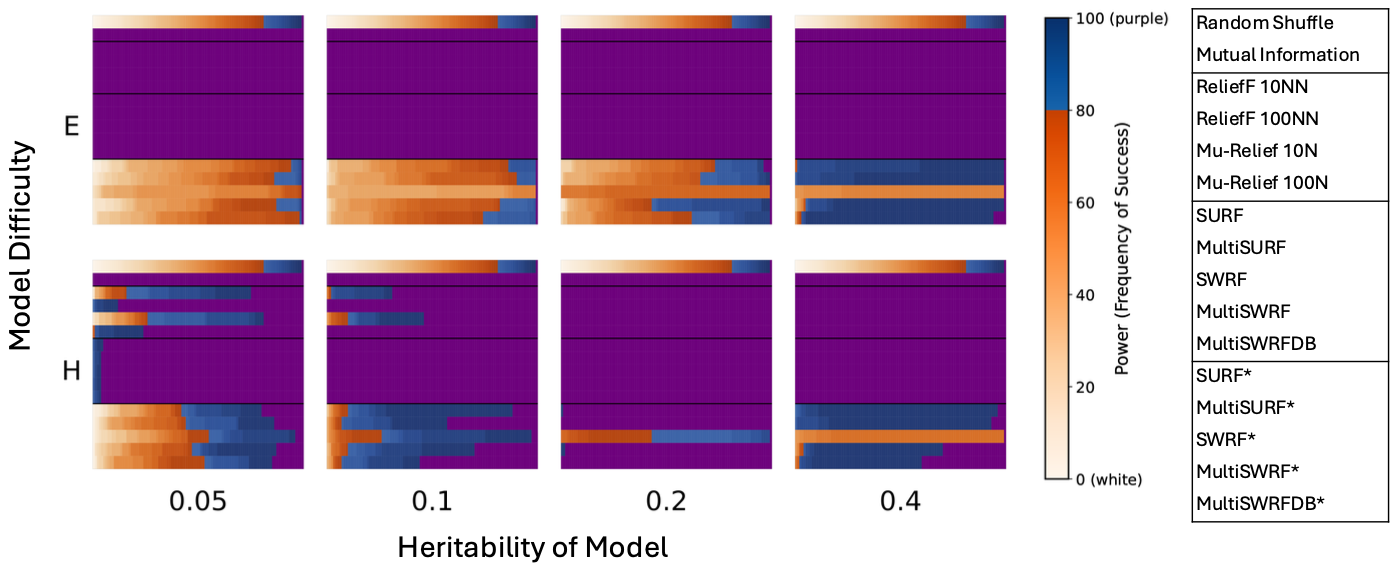}
\caption{Heatmap grid for the `1-Feature Main Effect' data group. X-axis indicates heritability of model values and y-axis indicates model difficulty: E (easy) or H (hard). All datasets have 1600 instances.}
\label{fig:1featuremaineff_unifiedheatmap}
\end{figure*}

The heatmap grid in Figure \ref{fig:1featuremaineff_unifiedheatmap} and Table S8 summarizes RBA performance across the 8 data configurations of the `1-Feature Main Effect' data group. The majority of `non-star' RBAs exhibited near-ideal performance across all dataset configurations (SURF, MultiSURF, SWRF, MultiSWRF, and MultiSWRFDB), with ReliefF (k=100) and $\mu$-Relief (k=100) slightly worse. Every `star' RBA performed significantly worse than all `non-star' RBAs (\textit{p} $< 0.0001$; see Supplementary File S5), reinforcing findings in \cite{Urbanowicz2018Benchmarking,Freda2024}. Notably, SWRF* performed least well of all RBAs, and all `star' algorithms performed similar to Random Shuffle on `Easy' datasets with the lowest heritability. Additionally, $\mu$-Relief (\emph{k}=10) significantly underperformed all `non-star' RBAs except ReliefF (\emph{k}=10) (\textit{p} $< 0.0001$). Likewise, ReliefF (\emph{k}=10) significantly underperformed all `non-star' RBAs except $\mu$-Relief (\emph{k}=10) (\textit{p} $< 0.001$). 

On the `2-Feature' and `4-Feature Additive Effect' data groups, including multiple predictive features with univariate effects, all `non-star' RBAs performed ideally while `star' RBAs were again less reliable ($p=0.054$ to $p<0.0001$; Supplementary Files S6 and S7) -- see Figures S3 and S4 \& Tables S9 and S10. Overall, RBAs that utilize more `near' neighbors in scoring, but no `far' neighbors, performed best across these univariate effect simulations.

\subsection{Scaling to Larger Feature Sets}

\begin{figure*}[!b]
\centering
\includegraphics[width=\textwidth]{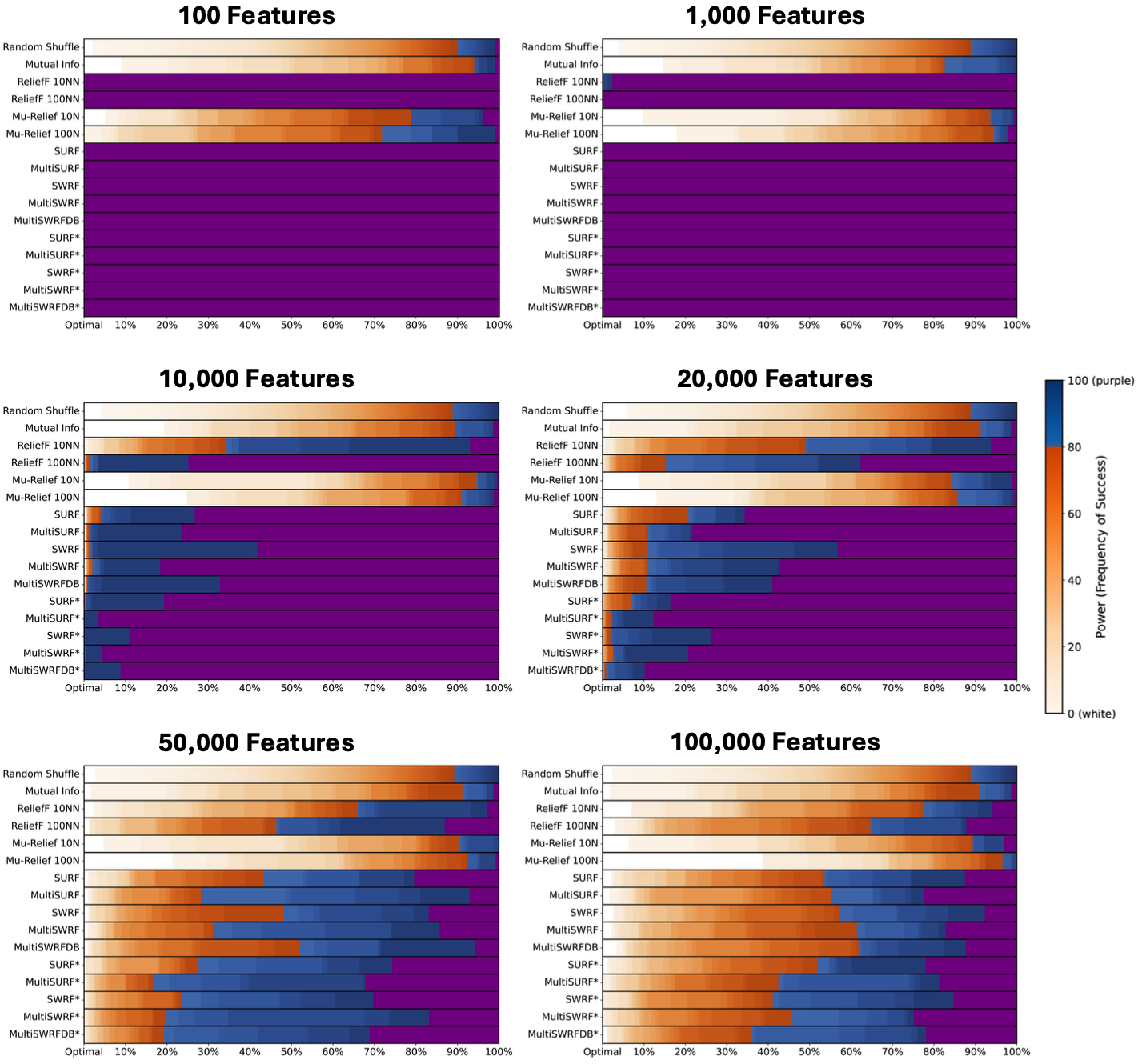}
\caption{Predictive feature ranking heatmaps for the `Number of Features (2-way Epistasis)' data group.}
\label{fig:largerfeatures2wayepi_unifiedheatmap}
\end{figure*}

 Figure \ref{fig:largerfeatures2wayepi_unifiedheatmap} summarizes RBA performance across most of the `Number of Features (2-way Epistasis)' data group. In contrast with previous benchmarking \cite{Urbanowicz2018Benchmarking,Freda2024}, this study examines 5 additional feature count levels (\textit{i.e.}, 2K, 5K, 8K, 20K, and 50K) towards better illustrating feature dimensionalities where `core' RBAs can reliably detect pure 2-way interactions. 
 
 The 100-feature configuration is the same as the top-right configuration in Figure \ref{fig:core2wayEpi_unifiedheatmap}, yielding ideal performance for all RBAs, except $\mu$-Relief (Table S11). The 1K-feature configuration yielded the same findings, but with SURF and ReliefF (\emph{k}=10) yielding non-significant performance losses (Table S12). The 10K-feature configuration departs from achieving any ideal RBA performance, but still yielded strong ranking performance for RBAs other than $\mu$-Relief. Here, `star' RBAs slightly but significantly outperform `non-star' RBAs ($p<0.0001$; Supplementary File S8) consistent with pure 2-way interaction trends in Figure \ref{fig:core2wayEpi_unifiedheatmap} and MultiSURF* standing out as best (Table S13). Also, RBAs that utilize the DB or gradient scoring showed a small advantage over SURF/SURF* ($p<0.01$ to $p<0.0001$ for SURF vs. `non-star' RBAs and $p=0.056$ to $p<0.05$ for SURF* vs. `star' RBAs). Furthermore,  ReliefF (\emph{k}=10) performance dropped considerably compared to other `non-star' RBAs ($p<0.0001$). 
 
 With greater than 10K features, RBA reliability in ranking predictive features involved in pure 2-way interactions progressively decreased, but `star' RBAs persisted as having better performance than `non-star' RBAs (Tables S14-S16). At 20K features, MultiSWRFDB* was best and consistently ranked predictive features in the top 10th percentile. At 50K features, MultiSURF* was best followed closely by MultiSWRFDB*. At 100K features, predictive feature ranking of most RBAs was still significantly better than Random Shuffle or Mutual Information and MultiSWRFDB* stood out as best. However, no RBA was able to consistently rank both predictive features higher than the 75th percentile. Furthermore, with 100K features, many (but not all) `star' RBAs slightly but significantly outperformed other `non-star' RBAs ($p < 0.05$ to $p < 0.0001$) and RBAs using the DB and/or gradient scoring yielded slightly better performance (non-significant; see Supplementary File S9).
 
While most RBA performance remained relatively strong at 10K features, we conducted followup analysis of detecting pure 2-way interactions at 2K, 5K, and 8K features (Figure S5, Tables S17-S19) and found that most `non-star' RBAs could maintain near-ideal ranking at 2K features, and most `star' RBAs could maintain near-ideal ranking at 8K features.

We also examined RBA scalability in the `Number of Features (Main Effect)' data group (Figure S6, Tables S20-S23). Every `non-star' RBA and Mutual Information achieved ideal performance at 100, 1K, 10K, and 100K features. However, the performance of every `star' RBA further declined as feature count increased. At and above 10K features, `star' RBAs performed no better than Random Shuffle. Here, SWRF* performed consistently worse than all other RBAs ($p < 0.0001$ for 10K and 100K features; Supplementary Files S10-S13).

\subsubsection{Runtime Improvements in \textit{scikit-rebate} }

RBAs achieved a performance speedup of 10-35x after \textit{scikit-rebate} code refactoring. Table S24 gives speedups achieved for previously benchmarked RBAs \cite{Urbanowicz2018Benchmarking}.

\subsection{Global Metrics for Different Scenarios}

Table \ref{tab:global_rankings} gives global mean/median metrics pooled over all primary datasets that are representative of 3 general scenarios influencing a practitioner's choice of feature selection strategy. These include detection of features involved in (1) 2-way epistasis as the only priority, (2) either 2-way epistasis and/or univariate effects, and (3) 2-way epistasis, univariate, and/or 3-way epistatic effects, as the most generalizable scenario. In Table \ref{tab:global_rankings}a (2-way epistasis only), `star' RBAs performed best in general, with MultiSWRFDB* performing best overall, and MultiSURF*, SWRF*, and MultiSWRF* performing nearly as well as MultiSWRFDB* (all with mean rankings within 0.33).

In Table \ref{tab:global_rankings}b (2-way + univariate); SWRF, MultiSWRF, MultiSURF, and MultiSWRFDB closely share best performance with mean rankings within 0.05. 

Towards identifying the most generalizable RBA, the algorithms ranked in Table \ref{tab:global_rankings}c (2-way + univariate + 3-way) are restricted to top performing RBAs from Table \ref{tab:global_rankings}b, \textit{i.e.}, the RBA has to be maximally capable on both 2-way epistasis and univariate effects (as the primary focus of this work). In this final scenario, MultiSWRFDB stood out as best due to its significantly better performance on 3-way epistasis datasets.

Of note, the top ranked RBA for each scenario, \textit{i.e.}, MultiSWRFDB*, SWRF, and MultiSWRFDB, were all newly proposed in this study. MultiSURF* and SWRF* ranked highly in (a), supporting previous benchmarking \cite{SWRFpaper,Urbanowicz2018Benchmarking,Freda2024}, and MultiSURF ranked highly for (b) and (c) scenarios, consistent with \cite{Urbanowicz2018Benchmarking,Freda2024}.

\begin{table*}[t]
\centering
\scriptsize
\setlength{\tabcolsep}{3pt}
\renewcommand{\arraystretch}{1.15}

\caption{Global RBA mean/median predictive feature ranking metric comparisons pooled by use scenario, \textit{i.e.}, groupings of `primary' data configurations. Each sub-table (a, b, and c) gives pooled metrics for the respective scenario. All rankings are on a 1-100 scale (1 = highest ranked feature, 100 = lowest ranked feature).}

\begin{subtable}[t]{0.27\textwidth}
\centering
\caption{2-way epistasis only}

\resizebox{\textwidth}{!}{
\begin{tabular}{lcc}
\toprule
RBA & Mean & Median \\
\midrule
\rowcolor{gray!15}\textbf{MultiSWRFDB*} & 9.92 & 2.00 \\
\rowcolor{gray!15}\textbf{MultiSURF*} & 10.11 & 2.00 \\
\rowcolor{gray!15}\textbf{SWRF*} & 10.19 & 2.00 \\
\rowcolor{gray!15}\textbf{MultiSWRF*} & 10.25 & 2.00 \\
SURF* & 11.10 & 2.00 \\
SWRF & 12.07 & 2.00 \\
MultiSWRF & 12.10 & 2.00 \\
MultiSURF & 12.13 & 2.00 \\
MultiSWRFDB & 12.15 & 2.00 \\
SURF & 12.95 & 2.73 \\
ReliefF 10NN & 16.08 & 3.00 \\
ReliefF 100NN & 18.40 & 3.00 \\
Mu-Relief 10N & 45.22 & 44.00 \\
Mu-Relief 100N & 48.98 & 48.00 \\
Mutual Info & 49.63 & 49.00 \\
Random Shuffle & 51.16 & 52.00 \\
\bottomrule
\end{tabular}
}
\end{subtable}
\hfill
\begin{subtable}[t]{0.341\textwidth}
\centering
\caption{2-way + Univariate}

\resizebox{\textwidth}{!}{
\begin{tabular}{lcc}
\toprule
RBA & Avg. Mean & Avg. Median \\
\midrule
\rowcolor{gray!15}\textbf{SWRF} & 6.75 & 1.50 \\
\rowcolor{gray!15}\textbf{MultiSWRF} & 6.77 & 1.50 \\
\rowcolor{gray!15}\textbf{MultiSURF} & 6.78 & 1.50 \\
\rowcolor{gray!15}\textbf{MultiSWRFDB} & 6.80 & 1.50 \\
SURF & 7.19 & 1.86 \\
ReliefF 10NN & 9.16 & 2.00 \\
ReliefF 100NN & 9.93 & 2.00 \\
MultiSWRFDB* & 14.69 & 4.00 \\
MultiSWRF* & 15.16 & 4.50 \\
MultiSURF* & 15.79 & 2.00 \\
SURF* & 16.62 & 2.50 \\
Mu-Relief 10N & 24.02 & 22.50 \\
Mu-Relief 100N & 25.24 & 24.50 \\
Mutual Info & 25.53 & 25.00 \\
SWRF* & 27.04 & 7.00 \\
Random Shuffle & 51.19 & 52.00 \\
\bottomrule
\end{tabular}
}
\end{subtable}
\hfill
\begin{subtable}[t]{0.33\textwidth}
\centering
\caption{2-way + Univariate + 3-way}

\resizebox{\textwidth}{!}{
\begin{tabular}{lcc}
\toprule
RBA & Avg. Mean & Avg. Median \\
\midrule
\rowcolor{gray!15}\textbf{MultiSWRFDB} & 9.20 & 2.33 \\
MultiSURF & 11.96 & 5.00 \\
SWRF & 13.87 & 8.00 \\
MultiSWRF & 14.88 & 10.33 \\
\bottomrule
\end{tabular}
}
\end{subtable}

\label{tab:global_rankings}
\end{table*}

\subsection{Secondary Dataset Performance}
In this study, secondary datasets serve as a sanity check for RBA applicability to other dataset types and scenarios. For most secondary dataset configurations (missing data, imbalanced class proportions, continuous-valued features, continuous-valued endpoint, or 3-class endpoint), most RBAs maintained ideal or near-ideal performance as in \cite{Urbanowicz2018Benchmarking} while $\mu$-Relief consistently displayed poor performance (not shown). However, on the 9-class configuration in the `Multi-Class Endpoint' data group, `star' RBAs completely failed, \textit{i.e.}, they regularly ranked predictive features with the lowest scores (Figure S7), which contradicted what was previously reported for MultiSURF* and SURF* in \cite{Urbanowicz2018Benchmarking} for the same dataset configuration.

\section{Discussion}

Selecting interacting features in high-dimensional genomic datasets remains a significant computational challenge. RBAs offer a tractable approach to this problem by maintaining sensitivity to complex epistatic interactions without prohibitive computational costs. This study benchmarked a comprehensive selection of `core' RBAs across diverse simulated genomic datasets to evaluate their feature ranking efficacy and identify the top performing RBAs. 

Results across dataset configurations reinforced previous findings that (1) nearly all RBAs were sensitive to detecting 2-way epistasis, (2) `star' RBAs, which use `far scoring', generally yielded a modest significant advantage in detecting 2-way epistasis over `non-star' RBAs, albeit at the expense of significant, often substantial performance losses in detecting univariate effects, (3) `non-star' RBAs offered balance between 2-way epistasis and univariate effect sensitivity, and (4) no RBA yielded reliable performance in detecting 4 or 5-way epistasis \cite{Greene2009SURF,MultiSURFpaper,SWRFpaper,Urbanowicz2018Benchmarking,Freda2024}. These findings held true for the 5 newly proposed RBAs which implemented novel combinations of key RBA elements (\textit{i.e.}, far scoring, recalculating neighborhood-defining metrics, DB zone, and gradient scoring).

Focusing on the latter two elements which both target the minimization of scoring updates from potentially less/non-informative middle-distance neighbors, we examine the utility of only employing a DB (MultiSURF/\allowbreak MultiSURF*), only gradient scoring (SWRF/\allowbreak SWRF*, MultiSWRF/\allowbreak MultiSWRF*), both (MultiSWRFDB/MultiSWR\discretionary{-}{}{}FDB*), or neither (SURF/\allowbreak SURF*). `Star' RBAs utilizing a DB and/or gradient achieved the highest sensitivity in 2-way epistasis datasets and their `non-star' counterparts were the strongest `non-star' RBAs on these datasets. While these elements had negligible impact on detecting univariate effects (Table S8), Table \ref{tab:global_rankings}b globally identified SWRF, MultiSWRF, MultiSWRFDB, and MultiSURF as comparably best, and Tables \ref{tab:global_rankings}a and \ref{tab:global_rankings}c ranked MultiSWRFDB* and MultiSWRFDB best, respectively. All these top-ranked RBAs employ a DB, gradient scoring, or both (for MultiSWRFDB* and MultiSWRFDB). Together, this suggests that minimizing the impact of `uninformative neighbors' located at `middle' distances from the target instance contributes to improving epistasis detection performance without harming main effect detection.

Focusing on whether recalculating neighborhood-defini\discretionary{-}{}{}ng metrics (\textit{i.e.}, `Multi'-RBAs) provided an advantage over global neighborhood-defining metrics, we compare performance between MultiSWRF/MultiSWRF* and SWRF/\allowbreak SWRF*. Results suggest a very limited, situation-specific impact, with negligible differences for 2-way epistasis detection (Table \ref{tab:global_rankings}a), but a slight improvement in MultiSWRF* over SWRF* on univariate effects (Table \ref{tab:global_rankings}b and Table S8), although neither performed strongly.

One notable observation was the failure of $\mu$-Relief in detecting 2-way epistasis and its slightly lower performance on univariate effects. This contradicts previous claims that $\mu$-Relief outperformed SURF*, MultiSURF, MultiSURF*, and ReliefF \cite{MuReliefpaper}. This highlights the importance of (1) first evaluating/benchmarking new algorithms within broader, carefully designed simulation studies where feature selection methods can be evaluated directly based on feature rankings rather than indirectly based on downstream modeling performance and (2) utilizing benchmarking datasets with sufficient interaction effects. Conversely, $\mu$-Relief performed best of all RBAs in detecting 3-way epistasis, suggesting opportunities for future work.

Other notable observations include: (1) `Star' RBA performance on univariate data (heritability = 0.4) became increasingly poor while `non-star' RBA performance remained consistent in larger feature sets (Figure S6), suggesting amplification of far-scoring's impact on univariate feature scores as described in \cite{Urbanowicz2018Benchmarking}; (2) SWRF* performed particularly poorly on univariate datasets, even compared to SURF* and MultiSWRF* which both also utilize every near and far `neighbor' in scoring (Figure \ref{fig:1featuremaineff_unifiedheatmap} and Table S8); and (3) `star' RBAs may have limited performance in data with many classes (Figure S7). 

Regarding RBA feature count scalability, this study identified 2K and 8K features as the maximum observed feature count within which `non-star' or `star' RBAs could ideally rank 2-way epistasis features, respectively. In practice, it's rarely necessary for feature selection to ideally rank predictive features, but rather just ensure they make the percentile cutoff applied to remove `irrelevant' features. The present results suggest that most `star' and `non-star' algorithms scale effectively up to 10K features with a minimum of $80\%$ power to rank both predictive features of a pure 2-way interaction above the 5th percentile (\textit{i.e.}, among the top 500/10,000 features), and expected $100\%$ power of all `star' RBAs above the 20th percentile (\textit{i.e.}, top 2000/10,000 features) (Figure \ref{fig:largerfeatures2wayepi_unifiedheatmap}). As previously reviewed \cite{Relief_review_paper}, a number of RBA-wrappers (\textit{e.g.}, TuRF \cite{moore2007tuning} and VLS \cite{eppstein2008very}) have been proposed that can be combined with any `core' RBA with the intention of improving predictive feature sensitivity within larger feature spaces. Characterization of RBA performance at different feature counts in this study is expected to offer useful empirical expectations on when and how to use such RBA-wrappers vs. when a `core' RBA alone may be sufficient to detect 2-way epistasis in a given dataset. At minimum, this study suggests using RBAs in combination with an RBA-wrapper in datasets with over 10K features. However, RBA-wrappers may add value even in datasets with over 5K-8K features. The runtime improvements introduced in this study are expected to make application of RBA-wrappers on high-dimensional datasets more computationally tractable.

Future work should address limitations of the present study; (1) expand evaluations to a number of diverse real-world datasets, ideally including some with known feature interactions, (2) examine a greater diversity of simulated 3-way interaction datasets, (3) expanded benchmarking on `secondary' dataset types, in particular with quantitative features and/or multi-class or quantitative endpoints, and compare benchmarked algorithms to those designed for continuous endpoints such as NPDR \cite{Le2020NPDR}, (4) examine the limits of RBA performance on even larger feature spaces (\textit{e.g.} 200K - 1 million features), in particular when combining top performing `core' RBAs with RBA-wrappers, and (5) consider strategies to extend and benchmark RBAs in \textit{scikit-rebate} to datasets with right-censored survival endpoints.

\section{Conclusion}
This study provides a comprehensive evaluation of the Relief-based algorithm (RBA) variants for interaction-sensitive feature selection. It (1) replicates and expands upon previous rigorous genomic simulation-study benchmarking, (2) evaluates and compares 12 RBAs (5 novel, 7 existing), (3) updates the \textit{scikit-rebate} RBA Python package; expanding it with 5 novel RBAs (MultiSWRFDB*, MultiSWRFDB, SWRF, MultiSWRF and MultiSWRF*) and 2 newly incorporated RBAs (SWRF* and $\mu$-Relief), additionally refactoring it yielding 10-35 fold speedups and adding precise discrimination of feature types, (4) examines the utility of RBA elements such as far-scoring, recalculating neighborhood metrics, a deadband zone, and gradient scoring, and (5) provides evidence-based RBA user recommendations for different scenarios - \textit{i.e.}, use `\textbf{MultiSWRFDB*}' for detecting 2-way epistasis only, use `\textbf{SWRF}' for detecting 2-way epistasis and/or univariate effects, use `\textbf{$\boldsymbol{\mu}$-Relief}' for 3-way epistasis only, and most generalizably use `\textbf{MultiSWRFDB}' for detecting all three. This work is expected to benefit anyone seeking to conduct feature selection; in particular, those working with noisy, high-dimensional datasets that potentially include complex multivariate associations as commonly found in biomedicine and various `omics' research.

\subsubsection*{CRediT authorship contribution statement}

\textbf{Kia Kazemi-Nia}: Writing -- original draft, Writing -- review \& editing, Software, Methodology, Visualization, Formal analysis, Validation, Data curation, Investigation. \textbf{Harsh Bandhey}: Software, Visualization, Data curation, Investigation. \textbf{Philip J. Freda}: Writing -- review \& editing, Supervision, Funding acquisition. \textbf{Ryan J. Urbanowicz}: Writing -- review \& editing, Methodology, Visualization, Data curation, Conceptualization, Project administration, Supervision, Funding acquisition.

\subsubsection*{Declaration of generative AI and AI-assisted technologies in the manuscript preparation process}

During the preparation of this work, the authors used ChatGPT to assist with code development. The authors reviewed and edited the output and take full responsibility for the content of the published article.

\subsubsection*{Declaration of competing interest}
The authors declare that they have no known competing financial interests or personal relationships that could have appeared to influence the work reported in this paper.

\subsubsection*{Data availability}
All simulated datasets used in this study are available upon request.

\subsubsection*{Acknowledgments}
The study was supported by Cedars Sinai Health Sciences University and NIH grants P30 AG073105, R01 AI173095, U01 AG066833, K01 DA063751, R01 HL175579, and P01 HL160471. Additional thanks to Tinghui Wu for contributions to \textit{scikit-rebate}.

\clearpage

\bibliographystyle{unsrtnat}
\bibliography{official}


\end{document}